\documentclass[runningheads]{llncs}

\usepackage{epsfig}
\usepackage{graphicx}
\usepackage{comment}
\usepackage{amsmath,amssymb} %
\usepackage{color}
\usepackage[export]{adjustbox}

\usepackage[font=small,skip=5pt]{caption}
\usepackage{float}

\newcommand{\specialcell}[2][c]{%
  \begin{tabular}[#1]{@{}c@{}}#2\end{tabular}}
\usepackage{multirow}
\usepackage{tabularx}
\usepackage{soul}
\usepackage{subfigure}
\usepackage{array}
\usepackage{comment}

\usepackage[mathscr]{euscript}
\newcolumntype{L}[1]{>{\raggedright\let\newline\\arraybackslash\hspace{0pt}}m{#1}}
\newcolumntype{C}[1]{>{\centering\let\newline\\arraybackslash\hspace{0pt}}m{#1}}
\newcolumntype{R}[1]{>{\raggedleft\let\newline\\arraybackslash\hspace{0pt}}m{#1}}
\newcolumntype{Y}{>{\centering\arraybackslash}X}
\usepackage{booktabs}
\usepackage{enumitem}

\usepackage[pagebackref=true,breaklinks=true,letterpaper=true,bookmarks=false]{hyperref}
\usepackage{amsmath}
\DeclareMathOperator*{\argmax}{arg\,max}

\renewcommand{\vec}[1]{\boldsymbol{#1}}
\newcommand{\mat}[1]{\mathbf{#1}}
\newcommand{\set}[1]{\mathcal{#1}}

\newcommand{\pose}[0]{\vec{\theta}}
\newcommand{\shape}[0]{\vec{\beta}}
\newcommand{\trans}[0]{\vec{t}}

\newcommand{\offsets}[0]{\mathbf{D}}

\newcommand{\pointset}[0]{\set{P}}
\newcommand{\point}[0]{\vec{p}}

\newcommand{\bodymesh}[0]{\set{B}}

\newcommand{\pcin}[0]{IP-Net}
\newcommand{\net}[3]{f^{#1}({#2}|{#3}w)}
\newcommand{\implicitdressed}[0]{\set{S}_o}
\newcommand{\implicitbody}[0]{\set{S}_{in}}
\newcommand{\implicitcorrespondence}[0]{I}

\newcommand{\myparagraph}[1]{\vspace{1pt}\noindent{\textbf{#1}}}
\graphicspath{ {./images/} }

\begin{document}
\pagestyle{headings}
\mainmatter
\def\ECCVSubNumber{3356}  %

\title{Combining Implicit Function Learning and Parametric Models for 3D Human Reconstruction} %

\titlerunning{IP-Net: Combining Implicit Functions and Parametric Modelling}
\author{Bharat Lal Bhatnagar\inst{1} \and
Cristian Sminchisescu\inst{2} \and
Christian Theobalt\inst{1} \and
Gerard Pons-Moll\inst{1}}
\authorrunning{BL. Bhatnagar et al.}
\institute{Max Planck Institute for Informatics, Saarland Informatics Campus\and Google Research\\
\email{\{bbhatnag, theobalt, gpons\}@mpi-inf.mpg.de, sminchisescu@google.com}}

\maketitle

\begin{abstract}
Implicit functions represented as deep learning approximations are powerful for reconstructing 3D surfaces. However, they can only produce static surfaces that are not controllable, which provides limited ability to modify the resulting model by editing its pose or shape parameters.
Nevertheless, such features are essential in building flexible models for both computer graphics and computer vision. In this work, we present methodology that combines detail-rich implicit functions and parametric representations in order to reconstruct 3D models of people that remain controllable and accurate even in the presence of clothing. Given sparse 3D point clouds sampled on the surface of a dressed person, we use an Implicit Part Network (\pcin{}) to jointly predict the \textit{outer} 3D surface of the dressed person, the \textit{inner} body surface, and the semantic correspondences to a parametric body model. We subsequently use correspondences to fit the body model to our inner surface and then non-rigidly deform it (under a parametric body + displacement model) to the outer surface in order to capture garment, face and hair detail.
In quantitative and qualitative experiments with both full body data and hand scans we show that the proposed methodology generalizes, and is effective even given incomplete point clouds collected from single-view depth images. Our models and code will be publicly released \footnote[1]{http://virtualhumans.mpi-inf.mpg.de/ipnet }.

\end{abstract}

\section{Introduction}
\begin{figure*}
\noindent\begin{minipage}[b]{0.48\textwidth}
    \includegraphics[width=\linewidth]{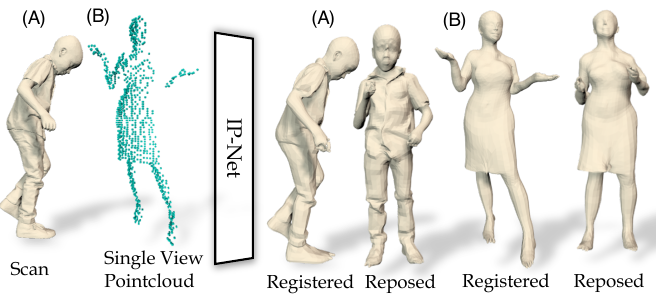}
    \caption{We combine implicit functions and parametric modeling for detailed and controllable reconstructions from sparse point clouds. \pcin{} predictions can be registered with SMPL+D model for control. \pcin{} can also register (A) 3D scans and (B) single view point clouds.}
\end{minipage}%
\hfill
\noindent\begin{minipage}[b]{0.48\textwidth}
    \includegraphics[width=\linewidth]{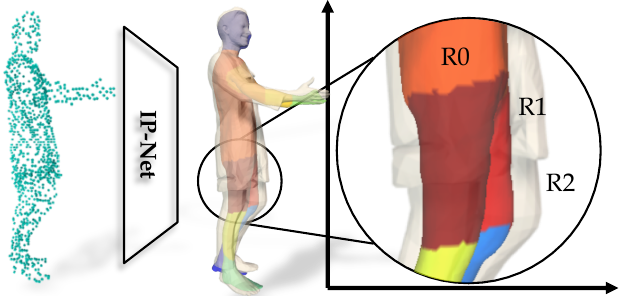}
    \caption{Unlike typical implicit reconstruction methods, \pcin{} predicts a double layered surface, classifying the points as lying inside the body (R0), between the body and the clothing (R1) and outside the clothing (R2). \pcin{} also  predicts part correspondences to the SMPL model.}
    \label{fig:teaser}
\end{minipage}
\end{figure*}

The sensing technology for capturing unstructured 3D point clouds is becoming ubiquitous and more accurate, thus opening avenues for extracting detailed models from point cloud data. This is important in many 3D applications such as shape analysis and retrieval, 3D content generation, 3D human reconstruction from depth data, as well as mesh registration, which is the workhorse of building statistical shape models~\cite{smpl2015loper,mh_joo2018total,xu2020ghum}. 
The problem is extremely challenging as the body can be occluded by clothing, hence identifying body parts given a point cloud is often ambiguous, and reasoning-with (or filling-in) missing data often requires non-local analysis. 
In this paper, we focus on the reconstruction of human models from sparse or incomplete point clouds, as captured by body scanners or depth cameras. In particular, we focus on extracting detailed 3D representations, including models of the underlying body shape and clothing, in order to make it possible to seamlessly re-pose and re-shape (\emph{control}) the resulting dressed human models. To avoid ambiguity, we refer to static implicit reconstructions as \textit{reconstruction} and our controllable model fit as \textit{registration}. Note that the \textit{registration} involves both reconstruction (explaining the given point cloud geometry) and  registration, as it is obtained by deforming a predefined model.

Learning-based methods are well suited to process sparse or incomplete point clouds,
as they can leverage prior data to fill in the missing information in the input,
but the choice of output representation limits either the resolution (when working with voxels or meshes), or the surface control (for implicit shape representations~\cite{i_DeepSDF,i_OccNet19,i_IMGAN19,ih_chibane2020}). 
The main limitation of learning an implicit function is that the output is ``just'' a static surface with \emph{no explicit model to control its pose and shape}. 
In contrast, parametric body models, such as SMPL~\cite{smpl2015loper} allow control, but the resulting meshes are overly-smooth and accurately regressing parameters directly from a point cloud is difficult (see Table~\ref{table:quantitaive-registration-comparison}). Furthermore, the surface of SMPL can not represent clothing, which makes registration difficult.
Non-rigidly registering a less constrained parametric model to point clouds using non-linear optimization is possible, but only yields good results when provided with very good initialization close to the data (without local assignment ambiguity) and the point cloud is reasonably complete (see Table~\ref{table:quantitaive-registration-comparison} and Fig.~\ref{fig:registration-comparison-point cloud-ifnet-scan-ours}).

The main idea in this paper is to take advantage of the best of both representations (implicit and parametric), and learn to predict body under clothing (including body part labels) in order to make subsequent optimization-based registration feasible. 
Specifically, we introduce a novel architecture which jointly learns 2 implicit functions for (i) the joint occupancy of the outer (body+clothing) and the inner (body) surfaces and (ii) body part labels. 
Following recent work \cite{ih_chibane2020}, we compute a 3-dimensional multi-scale tensor of deep features from the input point cloud, and make predictions at continuous query points.
Unlike recent work that only predicts the occupancy of a single surface~\cite{ih_chibane2020,i_DeepSDF,i_OccNet19,i_IMGAN19}, we jointly learn a continuous implicit function for the inner/outer surface prediction and another classifier for body part label prediction. 
Our key insight is that since the inner surface (body) can be well approximated by a parametric body model (SMPL), and the predicted body parts constrain the space of possible correspondences, fitting SMPL to the predicted inner surface is very robust. Starting from SMPL fitted to the inner surface, we register it to the outer surface (under an additional displacement model, SMLP+D~\cite{lazova3dv2019,alldieck2019learning}), which in turn allows us to \emph{re-pose and re-shape} the implicitly reconstructed outer surface. 

Our experiments show that our implicit network can accurately predict body shape under clothing, the outer surface, and part labels, which makes subsequent parametric model fitting robust.
Results on the Renderpeople dataset~\cite{renderpeople} demonstrate that our tandem of implicit function and parametric fitting yields detailed outer reconstructions, which are controllable, along with an estimation of body shape under clothing. 
We further achieve comparable performance on body shape under clothing on the BUFF dataset \cite{zhang2017shapeundercloth} without training on BUFF and without using temporal information.
To show that our model can be useful in other domains, we train it on the MANO dataset~\cite{MANO:SIGGRAPHASIA:2017} and show accurate registration using sparse and single view point clouds.
Our key contributions can be summarized as follows:
\begin{itemize}
    \item We propose a unified formulation which combines implicit functions and parametric modelling to obtain high quality controllable reconstructions from partial/ sparse/ dense point clouds of articulated dressed humans.
    \item Ours is the first approach to jointly reconstruct body shape under clothing along with full dressed reconstruction using a double surface implicit function, in addition to predicting part correspondences to a parametric model.
    \item Results on a dataset of articulated clothed humans and hands (MANO~\cite{MANO:SIGGRAPHASIA:2017}) show the wide applicability of our approach.
\end{itemize}

\section{Related Work}
In this section, we discuss works which extract 3D humans from visual observations using parametric and implicit surface models.
We further classify methods in top-down (optimization based) and bottom-up (learning based).

\subsection{Parametric Modelling for Humans}
Parametric body models factorize deformations into shape and pose~\cite{smpl2015loper,mh_joo2018total,xu2020ghum,Keyang_2020_ECCV}, soft-tissue~\cite{pons2015dyna}, and recently even clothing \cite{mh_bhatnagar2019mgn,mh_patel2020,tiwari20sizer}, which constraints meshes to the space of humans. 
Most of current \emph{model based approaches} optimize the pose and shape of SMPL~\cite{smpl2015loper} to match \emph{image features}, which are extracted with bottom-up predictors~\cite{mh_zanfir2018monocular,alldieck2018detailed,alldieck2018video,bogo2016smplify,mh_Xiang_2019_CVPR}. Alternative methods based on GHUM\cite{xu2020ghum} also exist\cite{zanfir2020weakly}.
The most popular image features are 2D joints~\cite{bogo2016smplify}, or 2D joints and silhouettes~\cite{alldieck2018video,alldieck2018detailed,mh_habermann}.
Some work have focused on estimating body shape under clothing~\cite{zhang2017shapeundercloth,bualan2008naked,yang2016estimation}, or capturing body shape and clothing jointly from scans~\cite{ponsmoll2017clothcap}. These approaches are typically slow, and are susceptible to local-minima.

In contrast, \emph{deep learning based} models predict body model parameters in a feed-forward pass~\cite{dibra2017human,mh_Rong_2019_ICCV,popa2017deep} and use bottom-up 2D features for self-supervision \cite{mh_kanazawa2018endtoend,mh_Pavlakos18,mh_Kolotouros19,mh_omran2018neural,mh_tung2017self,zanfir_nips2018} during training. 
These approaches are limited by the shape space of SMPL, can not capture clothing nor surface detail, and lack a feedback loop, which results in miss-alignments between reconstructions and input pixels. 

\emph{Hybrid methods} mitigate these problems by refining feed-forward predictions with optimization at training\cite{zanfir2020weakly} and/or test time\cite{zanfir2018monocular}, and by predicting displacements on top of SMPL, demonstrating capture of fine details and even \emph{clothing}~\cite{alldieck2019learning,mh_bhatnagar2019mgn}. However, the initial feed-forward predictions lack surface detail. 
Predicting normals and displacement maps on a UV-map or geometry image of the surface~\cite{mh_alldieck2019tex2shape,mh_Pumarola} results in more detail, but predictions are not necessarily aligned with the observations. 
 
Earlier work predicts dense correspondences on a depth map with a random forest and fit a 3D model to them~\cite{Pons-Moll_MRFIJCV,ponsmolMetricForests13,taylor2012vitruvian}.
To predict correspondences from point clouds using CNNs, depth maps can be generated where convolutions can be performed~\cite{wei2016dense}. 
Our approach differs critically in that i) we do not require generating multiple depth maps, 2) we predict the body shape under clothing which makes subsequent fitting easier, and
(ii) our approach can generate complete controllable and detailed surfaces from incomplete point clouds.

\subsection{Implicit Functions for Humans}
TSDFs~\cite{v_TSDF_CurlessL96} can represent the human surface implicitly, which is common in depth-fusion approaches~\cite{ih_Dynamic_Fusion,ih_KillingFusion}. 
Such free-form representation has been combined with SMPL~\cite{smpl2015loper} to increase robustness and tracking~\cite{ih_DoubleFusion}. Alternatively,
implicit functions can be parameterized with Gaussian primitives~\cite{mh_rhodin2016,mh_stoll}. Since these approaches are not learning based, they can not reconstruct the occluded part of the surface in single view settings. 

Voxels discretize the implicit occupancy function, which makes convolution operations possible. 
CNN based reconstructions using voxels~\cite{vh_varol2018bodynet,vh_Gilbert,vh_DeepHumans} or depth-maps~\cite{vh_Rogez,vh_facsimile,vh_Leroy18} typically 
produce more details than parametric models, but limbs are often missing. More importantly, unlike our method, the reconstruction quality is limited by the resolution of the voxel grid and increasing the resolution is hard as the memory footprint grows cubically.  

Recent methods learn a continuous implicit function representing the object surface directly~\cite{i_IMGAN19,i_DeepSDF,i_OccNet19}. However, these approaches have difficulties reconstructing articulated structures because they use a global shape code, and the networks tend to memorize typical object coordinates~\cite{ih_chibane2020}. 
The occupancy can be predicted based on local image features instead~\cite{ih_PiFu}, which results in medium-scale wrinkles and details, but the approach has difficulties with out of image plane poses, and is designed for image-reconstruction and can not handle point clouds. 
Recently, IF-Nets~\cite{ih_chibane2020} have been proposed for 3D reconstruction and completion from point clouds -- a mutliscale grid of deep features is first computed from the point cloud, and a decoder network classifies the occupancy based on mutli-scale deep features extracted at continuous point locations.  
These recent approaches~\cite{ih_chibane2020,ih_PiFu} make occupancy decisions based on local and global evidence, which results in more robust reconstruction of articulated and fine structures than decoding based on the X-Y-Z point coordinates and a global latent shape code~\cite{i_IMGAN19,i_DeepSDF,i_OccNet19,i_DeepLevelSet}.
However, they do not reconstruct shape under clothing and surfaces are not controllable. 
\subsection{Summary: Implicit vs Parametric Modelling}
Parametric models allow control over the surface and never miss body parts, but feed-forward prediction is hard, and reconstructions lack detail. 
Learning the implicit functions representing the surface directly is powerful because the output is continuous, details can be preserved better, and complex topologies can be represented. 
However, the output is not controllable, and can not guarantee that all body parts are reconstructed. Naive fitting of a body model to a reconstructed implicit surface often gets trapped into local minimal when the poses are difficult or clothing occludes the body (see Fig.~\ref{fig:registration-comparison-point cloud-ifnet-scan-ours}).  
These observations motivate the design of our hybrid method, which retains the benefits of both representations: i) control, ii) detail, iii) alignment with the input point clouds.

\section{Method}
\begin{figure*}[t]
\centering
    \includegraphics[width=\textwidth]{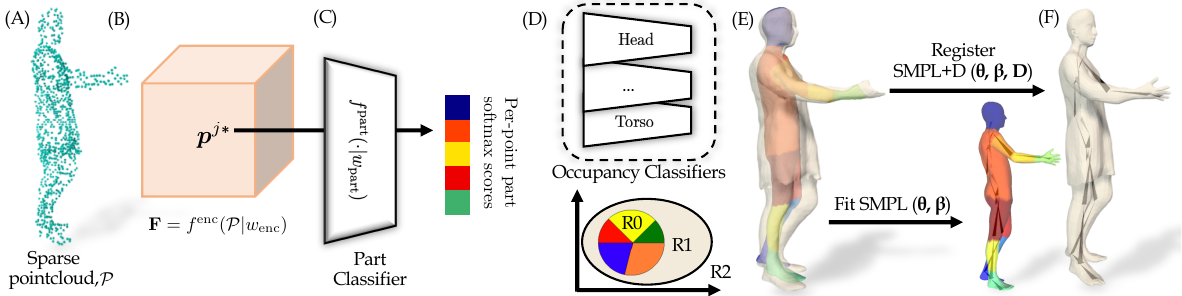}
\caption{The input to our method is (A) sparse point cloud $\pointset$. \pcin{} encoder $f^{\text{enc}(\cdot)}$ generates an (B) implicit representation of $\pointset$.
\pcin{} predicts, for each query point $\Vec{p}^j$, its (C) part label and double layered occupancy. \pcin{} uses (D) occupancy classifiers to classify the points as lying inside the body (R0), between the body and the clothing (R1) and outside the body (R2), hence predicting (E) full 3D shape $\implicitdressed$, body shape under clothing $\implicitbody$ and part labels.
We register \pcin{} predictions with (F) SMPL+D model to make implicit reconstruction controllable for the first time.}
 \label{fig:overview}
\end{figure*}

We introduce \pcin{}, a network to generate detailed 3D reconstruction from an unordered sparse point cloud. \pcin{} can additionally infer body shape under clothing and the body parts of the SMPL model. 
Training \pcin{} requires supervision on three fronts, i) an outer dressed surface occupancy--directly derived from 3D scans, ii) an inner body surface--we supervise with an optimization based body shape under clothing registration approach and iii) correspondences to the SMPL model--obtained by registering SMPL to scans using custom optimization.

\subsection{Training Data Preparation} 
\label{sec:data_prep}
To generate training data, we require non-rigidly registering SMPL~\cite{alldieck2019learning,lazova3dv2019} to 3D scans and estimating body shape under clothing, which is extremely challenging for the difficult poses in our dataset. Consequently, we first render the scans in multiple views, detect keypoints and joints, and integrate these as viewpoint landmark constraints to regularize registration similarly as in~\cite{alldieck2019learning,lazova3dv2019}. To non-rigidly deform SMPL to scans, we leverage SMPL+D~\cite{alldieck2019learning,lazova3dv2019}, which is an extension to SMPL that adds per-vertex free-form displacements on top of SMPL to model deformations due to garments and hair. For the body shape under clothing, we build on top of \cite{zhang2017shapeundercloth} and propose a similar optimization based approach integrating viewpoint landmarks. 
Once SMPL+D has been registered to the scans, we transfer body part labels from the SMPL model to the scans. We provide more details in the supplementary.
This process to generate training data is fairly robust, but required a lot of engineering to make it work. It also requires rendering multiple views of the scan, and does not work for sparse point clouds or scans without texture. 

One of the key contributions of this work is to replace this tedious process with \pcin{}, which quickly predicts a double layer implicit surface for body and outer surface, and body part labels to make subsequent registration using SMPL+D easy.  
We describe our network \pcin{}, that infers detailed geometry and SMPL body parts from sparse point clouds next.

\subsection{\pcin: Overview}
\label{sec:Network}
\pcin{} $\net{}{\cdot}{}$ takes in as input a sparse point cloud, $\pointset$ ($\sim$5k points), from articulated humans in diverse shapes, poses and clothing. \pcin{} learns an implicit function to jointly infer outer surface, $\implicitdressed$ (corresponding to full dressed 3D shape) and the inner surface $\implicitbody$ (corresponding to underlying body shape), of the person.
Since we intend to register SMPL model to our implicit predictions, \pcin{} additionally predicts, for each query point $\point^j \in \mathbb{R}^3$, the SMPL body part label $\implicitcorrespondence^j \in \{0, \hdots ,N-1\}$ (N=14) . We define $\implicitcorrespondence^j$ as a label denoting the associated body part on the SMPL mesh.

\myparagraph{\pcin{}: Feature encoding.} Recently, IF-Nets \cite{ih_chibane2020} achieve SOTA 3D mesh reconstruction from sparse point clouds. Their success can be attributed to two key insights: using a multi-scale, grid of deep features to represent shape, and predicting occupancy using features extracted at continuous point locations, instead of using the point coordinates. We build our \pcin{} encoder $f^{\text{enc}}(\cdot|w_\text{enc})$ in the spirit of IF-Net encoder. We denote our multi-scale grid-aligned feature representation as $\mat{F}=f^{\text{enc}}(\pointset|w_\text{enc})$ and the features at point $\point^j=(x,y,z)$ as $\mat{F}^j=\mat{F}(x,y,z)$.

\myparagraph{\pcin{}: Part classification.} Next, we train a multi-class classifier $f^{\text{part}}(\cdot|w_\text{part})$ that predicts, for each point $\point^j$, its part label (correspondence to nearest SMPL part) conditioned on its feature encoding. More specifically, $f^{\text{part}}(\cdot|w_\text{part})$ predicts a per part score vector $\Vec{D}^j \in [0,1]^N$ at every point $\point^j$
\begin{equation}
    \Vec{D}^j = f^{\text{part}}(\mat{F}^j|w_\text{part}).
\end{equation}
Then, we classify a point with the part label of maximum score

\begin{equation}
    \implicitcorrespondence^j = \argmax_{I \in \{0,\hdots,N-1 \}} (\Vec{D}_I^j).
\end{equation}
\myparagraph{\pcin{}: Occupancy prediction.} Previous implicit formulations \cite{i_OccNet19,i_DeepSDF,ih_PiFu,ih_chibane2020} train a deep neural network to classify points as being inside or outside a \emph{single} surface.
In addition, they minimize a classification/ regression loss over sampled points, which biases the network to perform better for parts with large surface area (more points) over smaller regions like hands (less points).

The key distinction between \pcin{} and previous implicit approaches is that it classifies points as belonging to 3 different regions: $0$-inside the body, $1$-between body and clothing and $2$-outside. This allows us to recover two surfaces (inner $\implicitbody$ and outer $\implicitdressed$), see Fig.~\ref{fig:teaser} and ~\ref{fig:overview}.  
Furthermore,  we use an ensemble of occupancy classifiers $\{f^I(\cdot|w_I)\}_{I=0}^{N-1}$, where each $f^I(\cdot|w_I):\mat{F}^j \mapsto \mathbf{o}^j \in [0,1]^3$ is trained to classify a point $\point^j$ with features $\mat{F}^j $ into the three regions $o^j \in \{0, 1, 2\}$, $o_j = \argmax_i \mathbf{o}^j_i $.
The idea here is to train the ensemble such that $f^I(\cdot|w_I)$ performs best for part $I$, 
and predict the final occupancy $o^j$ as a sum weighted by the part classification scores $\mat{D}_I^j\in \mathbb{R}$ at point $\vec{p}^j$
\begin{equation}
    o^j = \argmax_i \mathbf{o}^j_i,  \qquad \mathbf{o}^j = \sum_{I=0}^{N-1}\vec{D}^j_I \cdot f^{I}(\mat{F}^j|w_I),
\end{equation}
 thereby reducing the bias towards larger body parts. After dividing the space in 3 regions the double-layer surface is extracted from the two decision boundaries.

\myparagraph{\pcin{}: Losses} \pcin{} is trained using categorical cross entropy loss for both part-prediction ($f^{part}$) and occupancy prediction ($\{f^I\}_{I=0}^{N-1}$).

\myparagraph{\pcin{}: Surface generation} We use marching cubes \cite{LorensenC87mcubes} on our predicted occupancies to generate a triangulated mesh surface. \\
We provide more implementation details in the supplementary.

\subsection{Registering SMPL to \pcin{} Predictions}
\label{sec:SMPL_registration_implicit}
Implicit based approaches can generate details at arbitrary resolutions but reconstructions are static and not controllable. This makes these approaches unsuitable for re-shaping and re-posing. We propose the first approach to combine implicit reconstruction with parametric modelling which lifts the details from the implicit reconstruction onto the SMPL+D model \cite{alldieck2019learning,lazova3dv2019} to obtain an editable surface. We describe our registration using \pcin{} predictions next. 
We use SMPL to denote the parametric model constrained to undressed shapes, and SMPL+D (SMPL plus displacements) to represent details like clothing and hair. 

\myparagraph{Fit SMPL to implicit body:} We first optimize the SMPL shape, pose and translation parameters ($\pose, \shape, \trans$) to fit our inner surface prediction $\implicitbody$.
\begin{equation}
    E_{\text{data}}(\pose, \shape, \trans) =
    \frac{1}{|\implicitbody|}\sum_{\vec{v}_i \in \implicitbody}d(\vec{v}_i, \set{M}) +
    w \cdot \frac{1}{|\set{M}|}\sum_{\vec{v}_j \in \set{M}}d(\vec{v}_j, \implicitbody),
\end{equation}{}
where $\vec{v}_i$ and $\vec{v}_j$ denote vertices on $\implicitbody$ and SMPL surface $\set{M}$ respectively. $d(\vec{p}, \set{S})$ computes the distance of point $\vec{p}$ to surface $\set{S}$. In our experiments we set $w=0.1$

Additionally, we use the part labels predicted by \pcin{} to ensure that correct parts on the SMPL mesh explain the corresponding regions on the inner surface $\implicitbody$. This term is critical to ensure correct registration (see Table \ref{table:registration-comparison} and Fig. \ref{fig:registration-comparison})
\begin{equation}
    E_{\text{part}}(\pose, \shape, \trans) = \frac{1}{|\implicitbody|}\sum_{I=0}^{N-1} \sum_{\vec{v}_i \in \implicitbody} d(\vec{v}_i, \set{M}^I)\delta(\implicitcorrespondence^i=I),
\end{equation}{}
where $\set{M}^I$ denotes the surface of the SMPL mesh corresponding to part $I$ and $\implicitcorrespondence^i$ denotes the predicted part label of vertex $\vec{v}_i$.
The final objective can be written as follows
\begin{equation}
    E(\pose, \shape, \trans) = w_{\text{data}}E_{\text{data}} + w_{\text{part}}E_{\text{part}} + w_{\text{lap}}E_{\text{lap}},
\end{equation}{}
where $E_{\text{lap}}$ denotes a Laplacian regularizer. In our experiments we set the blancing weights $w_{\text{data/part/lap}}$ to 100, 10 and 1 respectively based on experimentation.

\myparagraph{Register SMPL+D to full implicit reconstruction:} Once we obtain the SMPL body parameters ($\pose, \shape, \trans$) from the above optimization, we jointly optimize the per-vertex displacements $\offsets$ to fit the outer implicit reconstruction $\implicitdressed$.
\begin{equation}
    E_{\text{data}}(\offsets, \pose, \shape, \trans) =
    \frac{1}{|\implicitdressed|}\sum_{\vec{v}_i \in \implicitdressed}d(\vec{v}_i, \set{M}) +
    w \cdot \frac{1}{|\set{M}|}\sum_{\vec{v}_j \in \set{M}}d(\vec{v}_j, \implicitdressed)
\end{equation}{}

\section{Dataset and Experiments}
\subsection{Dataset}
We train \pcin{} on a dataset of 700 scans \cite{twindom,treedys} and test on held out 50 scans \cite{renderpeople}.
We normalize our scans to a bounding box of size 1.6m.
To train \pcin{} we need paired data of sparse point clouds (input) and the corresponding outer surface, inner surface and correspondence to SMPL model (output). We generate the sparse point clouds by randomly sampling 5k points on our scans, which we voxelize into a grid of size 128x128x128 for our input.
We use the normalized scans directly as our ground truth dressed meshes and use our method for body shape registration under scan to get the corresponding body mesh $\bodymesh$ (see supplementary). For SMPL part correspondences,
we manually define 14 parts (left/right forearm, left/right mid-arm, left/right upper-arm, left/right upper leg, left/right mid leg, left/right foot, torso and head) on SMPL mesh and use the fact that our body mesh $\bodymesh$, is a template with SMPL-topology registered to the scan; this automatically annotates $\bodymesh$ with the part labels. The part label of each query point in $\mathbb{R}^3$, is the label of the nearest vertex on the corresponding body mesh $\bodymesh$. Note that part annotations do not require manual effort.

We evaluate the implicit outer surface reconstructions against the GT scans. We use the optimization based approach described in Sec. \ref{sec:data_prep} to obtain ground truth registrations.
\begin{table}[t]
\begin{minipage}[t!]{0.52\linewidth}
\begin{center}
\begin{tabularx}{\textwidth}{l *{2}{Y}}
\toprule[1.5pt]
\specialcell{Register} &  \specialcell{\bf Outer} &  \specialcell{\bf Inner} \\
\specialcell{SMPL+D} &  \specialcell{\bf reg.} &  \specialcell{\bf reg.} \\
\midrule
(a) Sparse point cloud & 14.85 & NP* \\
(b) IF-Net \cite{ih_chibane2020} & 13.88 & NP* \\
(c) Regress SMPL+D params & 32.45 & NP* \\
(d) {\bf \pcin{} (Ours)} & {\bf 3.67} & {\bf 3.32} \\
\bottomrule[1.5pt]
\end{tabularx}
\caption{\pcin{} predictions, i.e. the outer/ inner surface and correspondences to SMPL are key to high quality SMPL+D registration. We compare the quality (vertex-to-vertex error in cm) of registering to (a) point cloud, (b) implicit reconstruction by IF-Net\cite{ih_chibane2020}, (c) regressing SMPL+D params and (d) \pcin{} predictions. NP* means `not possible'.}
\label{table:quantitaive-registration-comparison}
\end{center}
\end{minipage} \;
\begin{minipage}[t!]{0.47\linewidth}
\begin{center}
\begin{tabular}{lcc}
\toprule[1.5pt]
&  \specialcell{\bf Outer} & \specialcell{\bf Inner}\\
\midrule
(a) outer only & 11.84 & 11.62\\
(b) outer+inner & 11.54 & 11.14\\
(c) {\bf outer+inner+parts} & \bf{3.67} & \bf{3.32}\\
\bottomrule[1.5pt]
\end{tabular}
\caption{We compare three possibilities of registering the SMPL model to the implicit reconstruction produced by \pcin. (a) registering SMPL+D to outer implicit reconstruction, (b) registering SMPL+D using the body prediction and (c) registering SMPL+D using body and part predictions. We report vertex-to-vertex error (cm) between the GT and predicted registered meshes.}
\label{table:registration-comparison}
\end{center}
\end{minipage}%
\end{table}

\begin{figure*}[t]
    \includegraphics[width=1\textwidth]{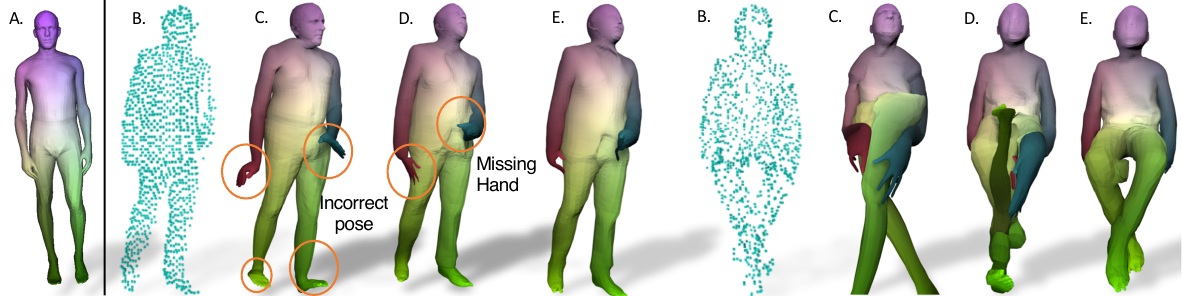}
\caption{We compare quality of SMPL+D registration for various alternatives to \pcin{}. We show A) colour coded reference SMPL, B) the input point cloud, C) registration directly to sparse PC, D) registration to IFNet \cite{ih_chibane2020} prediction and E) registration to \pcin{} predictions. It is important to note that poses such as sitting (second set) are difficult to register without explicit correspondences to the SMPL model.}
 \label{fig:registration-comparison-point cloud-ifnet-scan-ours}
\end{figure*}

\begin{table}[t]
\begin{minipage}[t!]{0.5\linewidth}
\begin{center}
\begin{tabularx}{\textwidth}{l *{2}{Y}}
\toprule[1.5pt]
\specialcell{Register, single} &  \specialcell{\bf Outer} &  \specialcell{\bf Inner} \\
\specialcell{view point cloud} &  \specialcell{\bf reg.} &  \specialcell{\bf reg.} \\
\midrule
Sin. view PC & 15.90 & NP* \\
Sin. view PC + \pcin{} & & \\
correspondences (Ours) & 14.43 & NP* \\
{\bf \pcin{} (Ours)} & {\bf 5.11} & {\bf 4.67} \\
\bottomrule[1.5pt]
\end{tabularx}
\caption{Depth sensors can provide single depth view point clouds. We report registration accuracy (vertex-to-vertex distance in cm) on such data and show that registration using \pcin{} predictions
is significantly better than alternatives. NP* implies `not possible'.}
\label{table:quantitaive-registration-comparison-SV}
\end{center}
\end{minipage} \;
\hfill
\begin{minipage}[t!]{0.5\linewidth}
\begin{center}
\begin{tabularx}{\textwidth}{l *{2}{Y}}
\toprule[1.5pt]
\specialcell{Register {\bf with \pcin{}}} &  \specialcell{\bf Outer} &  \specialcell{\bf Inner} \\
\specialcell{correspondences} &  \specialcell{\bf reg.} &  \specialcell{\bf reg.} \\
\midrule
Sparse point cloud & 13.93 & NP* \\
Scan & 3.99 & NP* \\
\bf{\pcin{} (Ours)} & {\bf 3.67} & {\bf 3.32} \\
\bottomrule[1.5pt]
\end{tabularx}
\caption{An interesting use for \pcin{} is to fit the SMPL+D model to sparse point clouds or scans using its part labels. This is useful for scan registration as we can retain the details of the high resolution scan and make it controlable. We report vertex-to-vertex error in cm. See Fig. \ref{fig:registration-comparison-scan} for qualitative results. NP* implies `not possible'.}
\label{table:quantitaive-registration-comparison-with-correspondences}
\end{center}
\end{minipage}%
\end{table}

\begin{figure*}[t]
\centering
    \includegraphics[width=1\textwidth]{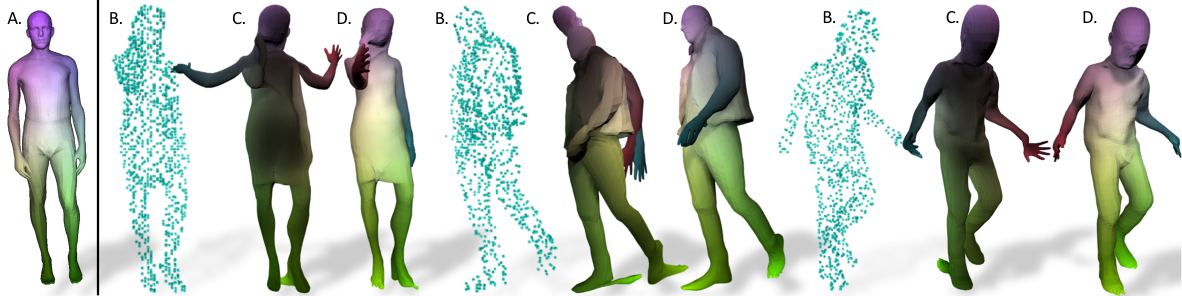}
\caption{We highlight the importance of \pcin{} predicted correspondences for accurate registration. We show A) color coded SMPL vertices to appreciate registration quality and three sets of comparative results. In each set, we visualize B) the input point cloud, C) registration without using \pcin{} correspondences and D) registration with \pcin{} correspondences. It can be seen that without correspondences we find problems like 180$^{\circ}$ flips (dark colors indicate back surface), vertices from torso being used to explain arms etc.
These results are quantitatively corroborated in Table \ref{table:registration-comparison}.}
\label{fig:registration-comparison}
\end{figure*}

\begin{figure*}[t]
    \includegraphics[width=1\textwidth]{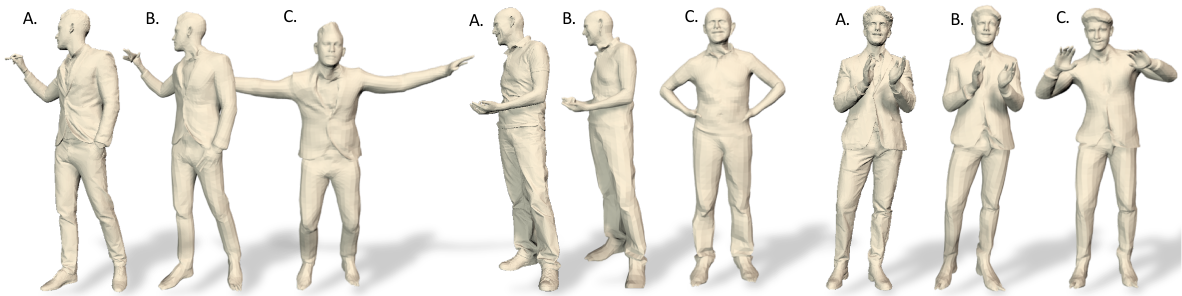}
\caption{\pcin{} can be used for scan registration. As can be seen from Table \ref{table:quantitaive-registration-comparison}, registering SMPL+D directly to scan is difficult. We propose to predict the inner body surface and part correspondences for every point on the scan using \pcin{} and subsequently register SMPL+D to it. This allows us to retain outer geometric details from the scan while also being able to animate it. We show A) input scan, B) SMPL+D registration using \pcin{}, C) scan in a novel pose. See video at \cite{ipnet}.}
 \label{fig:registration-comparison-scan}
\end{figure*}

\begin{figure*}[t]
    \includegraphics[width=1\textwidth]{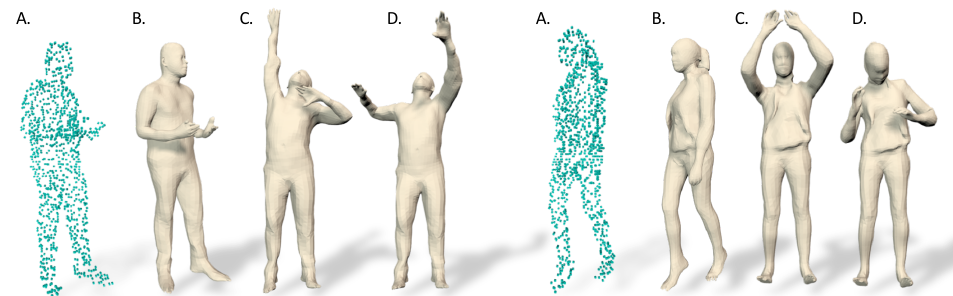}
\caption{Implicit predictions by \pcin{} can be registered with SMPL+D model and hence reposed. We show, A) input point cloud, B) corresponding SMPL+D registration and C,D) two instances of new poses.}
 \label{fig:reposed_predictions}
\end{figure*}

\begin{figure*}[t]
\centering
    \includegraphics[width=1\textwidth]{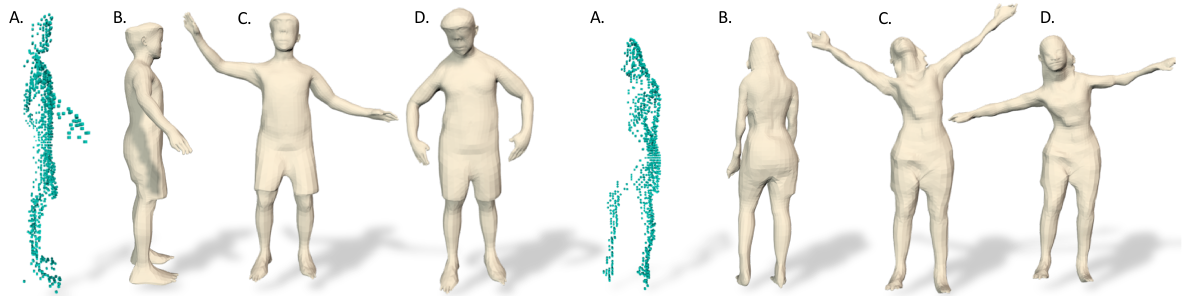}
\caption{Single depth view point clouds (A) are becoming increasingly accessible with devices like Kinect. We show our registration using \pcin{} (B) and reposing results (C,D) with two novel poses using such data.}
 \label{fig:single-view}
\end{figure*}

\begin{figure*}[t]
\centering
    \includegraphics[width=1\textwidth]{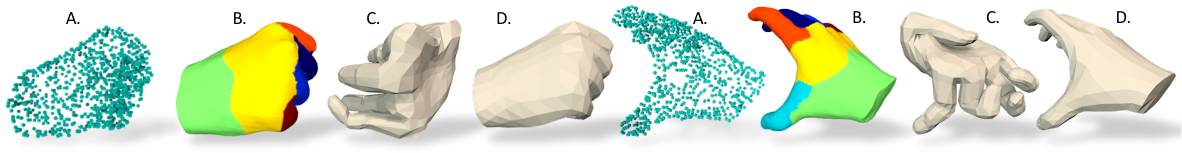}
    \includegraphics[width=1\textwidth]{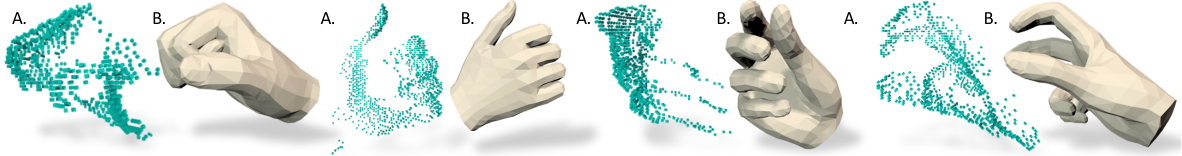}
\caption{We extend our idea of predicting implicit correspondences to parametric models to 3D hands.
Here, we show results on MANO hand dataset \cite{MANO:SIGGRAPHASIA:2017}. In the first row we show A) input PC, B) surface and part labels predicted by \pcin{}, C) registration without part correspondences, and D) our registration. Registration without part labels is ill-posed and often leads to wrong parts explaining the surface.
In the second row we show A) input single-view PC and B) corresponding registrations using \pcin{}.}
 \label{fig:MANO}
\end{figure*}

\subsection{Outer surface reconstruction.}
For the task of outer surface reconstruction, we demonstrate that \pcin{} performs better or on par with state of the art implicit reconstruction methods, Occ.Net \cite{i_OccNet19} and IF-Net \cite{ih_chibane2020}.
We report the average bi-directional vertex-to-surface error of 9.86mm, 4.86mm and 4.95mm for \cite{i_OccNet19},  \cite{ih_chibane2020} and \pcin{} respectively. We show qualitative results in the supplementary. Unlike \cite{i_OccNet19,ih_chibane2020} which predict only the outer surface, we infer body shape under clothing and body part labels with the same model.

\subsection{Comparison to Baselines}
The main contribution of our method is to make implicit reconstructions controllable.
We do so by registering SMPL+D model \cite{alldieck2018video,lazova3dv2019} to \pcin{} outputs: outer surface, inner surface and part correspondences.
This raises the the following questions, ``Why not a) register SMPL+D directly to the input sparse point cloud?, b) register SMPL+D to the surface generated by an existing reconstruction approach \cite{ih_chibane2020}? c) directly regress SMPL+D parameters from the point cloud? and d) How much better is it to register using \pcin{} predictions?''. Table \ref{table:quantitaive-registration-comparison} and Fig. \ref{fig:registration-comparison-point cloud-ifnet-scan-ours} show that option d) (our method) is significantly better than the other baselines (a,b and c).
To regress SMPL+D parameters (Option c), we implement a feed forward network that uses a similar encoder as \pcin{}, but instead of predicting occupancy and part labels, produces SMPL+D parameters. We notice that the error for this method is dominated by misaligned pose and overall scale of the prediction. If we optimise the global orientation and scale of the predictions, this error is reduced from 32.45cm to 7.25cm which is still very high as compared to \pcin{} based registration (3.67cm) which requires no such adjustments.
This experiment provides two key insights, i) it is significantly better to make local predictions using implicit functions and later register a parametric model, than to directly regress the parameters of the model and ii) directly registering a parametric model to an existing reconstruction method~\cite{ih_chibane2020} yields larger errors than registering to \pcin{} outputs (13.88cm vs 3.67cm). 

\subsection{Body Shape under Clothing} \label{Buff_exp}
We quantitatively evaluate our body shape predictions on BUFF dataset \cite{zhang2017shapeundercloth}. Given a sparse point cloud generated from BUFF scans, \pcin{} predicts the inner and outer surfaces along with the correspondences. We use our registration approach, as described in Sec. \ref{sec:SMPL_registration_implicit} to fit SMPL to our inner surface prediction and evaluate the error as per the protocol described in \cite{zhang2017shapeundercloth}.
It is important to note that the comparison is unfair to our approach on several counts:
\begin{enumerate}
    \item Our network uses sparse point clouds whereas \cite{zhang2017shapeundercloth} use 4D scans for their optimization based approach.
    \item Our network was not trained on BUFF (noisier scans, missing soles in feet).
    \item The numbers reported by \cite{zhang2017shapeundercloth} are obtained by jointly optimizing the body shape over entire 4D sequence, whereas our network makes a per-frame prediction without using temporal information.
\end{enumerate}
We also compare our method to \cite{yang2016estimation}. We report the following errors (mm): (\cite{zhang2017shapeundercloth} male: 2.65, female: 2.48), (\cite{yang2016estimation} male: 17.85, female: 18.19) and (Ours male: 3.80, female: 6.17).
Note that we did not have gender annotations for training \pcin{} and hence generated our training data by registering all the scans to the `male' SMPL model. This leads to significantly higher errors in estimating the body shape under clothing for `female' subjects (we think this could be fixed by fitting gender specific models during training data generation). We add subject and sequence wise comparison in the supplementary. We show that our approach can accurately infer body shape under clothing using just a sparse point cloud and is on par with approaches which use much more information.

\subsection{Why is correspondence prediction important?}
In this experiment, we demonstrate that inner surface reconstruction and part correspondences predicted by \pcin{} are key for accurate registration.
We discuss three obvious approaches for this registration:
\begin{enumerate}[label=(\alph*)]
    \item Register SMPL+D directly to the implicit outer surface predicted by \pcin{}. This approach is simple and can be used with any other existing implicit reconstruction approaches.
    \item Register SMPL to the inner surface predicted by \pcin{} and then non-rigidly register to the outer surface (without leveraging the correspondences).
    \item (Ours) First fit the SMPL model to the inner surface using correspondences and then non-rigidly register SMPL+D model to the implicit outer surface.
\end{enumerate}
We report our results for the aforementioned approaches in Table \ref{table:registration-comparison} and Fig. \ref{fig:registration-comparison}. It can clearly be seen (Fig. \ref{fig:registration-comparison}, first set) that the arms of the SMPL model have not snapped to the correct pose. This is to be expected when arms are close to the body and no joint or correspondence information is present. In the second set, we see that vertices from torso are being used to explain the arms while SMPL arms are left hanging out. Third set is the classic case of 180$^\circ$ flipped fitting (dark color indicates back surface). This experiment highlights the importance of inner body surface and part correspondence prediction.\\

\subsection{Why not independent networks for inner \& outer surfaces?}
\pcin{} jointly predicts the inner and the outer surface for a human with clothing. Alternatively, one could train two separate implicit reconstruction networks using an existing SOTA approach. This has a clear disadvantage that one surface cannot reason about another, leading to severe inter-penetrations between the two. We report the average surface area of intersecting mesh faces which is 2000.71mm$^2$ for the two independent network approach, whereas with \pcin{} the number is 0.65mm$^2$, which is four orders of magnitude smaller. We add qualitative results in the supplementary. Our experiment demonstrates that having a joint model for inner and outer surfaces is better.

\subsection{Using \pcin{} Correspondences to Register Scans}
A very powerful use case for \pcin{} is scan registration. Current state-of-the art registration approaches \cite{alldieck2019learning,lazova3dv2019} for registering SMPL+D to 3D scans are tedious and cumbersome (as described in Sec. \ref{sec:data_prep}).
We provide a simple alternative using \pcin{}. We sample points on our scan and generate the voxel grid used by \pcin{} as input. We then run our pre-trained network and estimate the inner surface corresponding to the body shape under clothing. We additionally predict correspondences to the 
SMPL model for
\emph{each vertex on the scan}. We then use our registration (Sec. \ref{sec:SMPL_registration_implicit}) to fit SMPL to the inner surface and then non-rigidly register SMPL+D to the scan surface, hence replacing the requirement for accurate 3D joints with \pcin{} part correspondences. We show the scan registration and reposing results in Fig. \ref{fig:registration-comparison-scan} and Table \ref{table:quantitaive-registration-comparison-with-correspondences}. This is a useful experiment that shows that feed-forward \pcin{} predictions can be used to replace tedious bottlenecks in scan registration.

\subsection{Registration From Point Clouds Obtained from a Single View}
We show that \pcin{} can be trained to process sparse point clouds from a single view (such as from Kinect). We show qualitative and quantitative results in Fig. \ref{fig:single-view} and Table \ref{table:quantitaive-registration-comparison-SV}, which demonstrate that IP-Net predictions are crucial for successful fitting in this difficult setting. This experiment highlights the general applicability of \pcin{} to a variety of input modalities ranging from dense point clouds such as scans to sparse point clouds to single view point clouds.

\subsection{Hand Registration}
We show the wide applicability of IP-Net by using it for hand registration.
We train \pcin{} on the MANO hand dataset \cite{MANO:SIGGRAPHASIA:2017} and show hand registrations to full and single view point cloud in Fig. \ref{fig:MANO}. We report an avg. vertex-to-vertex error of 4.80mm and 4.87mm in registration for full and single view point cloud respectively. This experiment shows that the idea of predicting implicit correspondences to a parametric model can be generically applied to different domains.

\myparagraph{Limitations of IP-Net.}
During our experiments we found IP-Net does not perform well with poses that were very different than the training set. We also feel that the reconstructed details can be further improved especially around the face. We encourage the readers to see supplementary for further discussion.

\section{Conclusions}

Learning implicit functions to model humans has been shown to be powerful but the resulting representations are not amenable to control or reposing which are essential for both animation and inference in computer vision. We have presented methodology to combine expressive implicit function representations and parametric body modelling in order to produce 3D reconstructions of humans that remain controllable even in the presence of clothing. 

Given a sparse point cloud representing a human body scan, we use implicit representation obtained using deep learning in order to jointly predict the \textit{outer} 3D surface of the dressed person and the \textit{inner} body surface as well as the semantic body parts of the parametric model. We use the part labels to fit the parametric model to our inner surface and then non-rigidly deform it (under a body prior + displacement model) to the outer surface in order to capture garment, face and hair details.
Our experiments demonstrate that 1) predicting a double layer surface is useful for subsequent model fitting resulting in reconstruction improvements of $3$mm and 2) leveraging semantic body parts is \emph{crucial} for subsequent fitting and results in improvements of $8.17$cm. 
The benefits of our method are paramount for difficult poses or when input is incomplete such as single view sparse point clouds, where the double layer implicit reconstruction and part classification is essential for successful registration.  
Our method generalizes well to other domains such as 3D hands (as evaluated on the MANO dataset) and even works well when presented with incomplete point clouds from a single depth view, as shown in extensive quantitative and qualitative experiments.

\myparagraph{Acknowledgements.} We thank Neng Qian, Jiayi Wang and Franziska Mueller for help with MANO experiments, Tribhuvanesh Orekondy for discussions and the reviewers for their feedback. Special thanks to RVH team members \cite{rvh_grp}, their feedback significantly improved the overall writing and readability of this manuscript.
We thank Twindom \cite{twindom} for providing data for this project.
This work is funded by the Deutsche Forschungsgemeinschaft (DFG, German Research Foundation) - 409792180 (Emmy Noether Programme,
project: Real Virtual Humans), ERC Consolidator Grant 4DRepLy (770784) and Google Faculty Research Award.

\par\vfill\par

\clearpage
\bibliographystyle{splncs04}
\bibliography{egbib}

\begin{thebibliography}{10}
\providecommand{\url}[1]{\texttt{#1}}
\providecommand{\urlprefix}{URL }
\providecommand{\doi}[1]{https://doi.org/#1}

\bibitem{renderpeople}
https://renderpeople.com/

\bibitem{twindom}
https://web.twindom.com/

\bibitem{treedys}
https://www.treedys.com/

\bibitem{ipnet}
http://virtualhumans.mpi-inf.mpg.de/ipnet

\bibitem{rvh_grp}
http://virtualhumans.mpi-inf.mpg.de/people.html

\bibitem{alldieck2019learning}
Alldieck, T., Magnor, M., Bhatnagar, B.L., Theobalt, C., Pons-Moll, G.:
  Learning to reconstruct people in clothing from a single {RGB} camera. In:
  {IEEE}/{CVF} Conference on Computer Vision and Pattern Recognition ({CVPR})
  (2019)

\bibitem{alldieck2018detailed}
Alldieck, T., Magnor, M., Xu, W., Theobalt, C., Pons-Moll, G.: Detailed human
  avatars from monocular video. In: International Conf. on 3D Vision (sep 2018)

\bibitem{alldieck2018video}
Alldieck, T., Magnor, M., Xu, W., Theobalt, C., Pons-Moll, G.: Video based
  reconstruction of {3D} people models. In: {IEEE} Conf. on Computer Vision and
  Pattern Recognition (2018)

\bibitem{mh_alldieck2019tex2shape}
Alldieck, T., Pons-Moll, G., Theobalt, C., Magnor, M.: Tex2shape: Detailed full
  human body geometry from a single image. In: {IEEE} International Conference
  on Computer Vision ({ICCV}). {IEEE} (oct 2019)

\bibitem{bualan2008naked}
B{\u{a}}lan, A.O., Black, M.J.: The naked truth: Estimating body shape under
  clothing. In: Forsyth, D., Torr, P., Zisserman, A. (eds.) European Conf. on
  Computer Vision. pp. 15--29. Springer, Springer Berlin Heidelberg (2008)

\bibitem{mh_bhatnagar2019mgn}
Bhatnagar, B.L., Tiwari, G., Theobalt, C., Pons-Moll, G.: Multi-garment net:
  Learning to dress 3d people from images. In: {IEEE} International Conference
  on Computer Vision ({ICCV}). {IEEE} (oct 2019)

\bibitem{bogo2016smplify}
Bogo, F., Kanazawa, A., Lassner, C., Gehler, P., Romero, J., Black, M.J.: Keep
  it {SMPL}: Automatic estimation of {3D} human pose and shape from a single
  image. In: Leibe, B., Matas, J., Sebe, N., Welling, M. (eds.) European Conf.
  on Computer Vision. Springer International Publishing (2016)

\bibitem{i_IMGAN19}
Chen, Z., Zhang, H.: Learning implicit fields for generative shape modeling.
  In: {IEEE} Conference on Computer Vision and Pattern Recognition, {CVPR}
  2019, Long Beach, CA, USA, June 16-20, 2019. pp. 5939--5948 (2019)

\bibitem{ih_chibane2020}
Chibane, J., Alldieck, T., Pons-Moll, G.: Implicit functions in feature space
  for 3d shape reconstruction and completion. In: {IEEE} Conference on Computer
  Vision and Pattern Recognition (CVPR). {IEEE} (Jun 2020)

\bibitem{v_TSDF_CurlessL96}
Curless, B., Levoy, M.: A volumetric method for building complex models from
  range images. In: Proceedings of the 23rd Annual Conference on Computer
  Graphics and Interactive Techniques, {SIGGRAPH} 1996, New Orleans, LA, USA,
  August 4-9, 1996. pp. 303--312. Association for Computing Machinery, New
  York, NY, United States (1996)

\bibitem{dibra2017human}
Dibra, E., Jain, H., Oztireli, C., Ziegler, R., Gross, M.: Human shape from
  silhouettes using generative hks descriptors and cross-modal neural networks.
  In: {IEEE} Conf. on Computer Vision and Pattern Recognition (2017)

\bibitem{vh_Rogez}
Gabeur, V., Franco, J., Martin, X., Schmid, C., Rogez, G.: Moulding humans:
  Non-parametric 3d human shape estimation from single images. {IEEE}
  International Conference on Computer Vision, {ICCV}  (2019)

\bibitem{vh_Gilbert}
Gilbert, A., Volino, M., Collomosse, J.P., Hilton, A.: Volumetric performance
  capture from minimal camera viewpoints. In: Computer Vision - {ECCV} 2018 -
  15th European Conference, Munich, Germany, September 8-14, 2018, Proceedings,
  Part {XI}. pp. 591--607 (2018). \doi{10.1007/978-3-030-01252-6\_35}

\bibitem{mh_habermann}
Habermann, M., Xu, W., Zollh\"{o}fer, M., Pons-Moll, G., Theobalt, C.: Livecap:
  Real-time human performance capture from monocular video. ACM Trans. Graph.
  \textbf{38}(2),  14:1--14:17 (Mar 2019). \doi{10.1145/3311970}

\bibitem{mh_joo2018total}
Joo, H., Simon, T., Sheikh, Y.: Total capture: A 3d deformation model for
  tracking faces, hands, and bodies. In: Proceedings of the IEEE conference on
  computer vision and pattern recognition. pp. 8320--8329 (2018)

\bibitem{mh_kanazawa2018endtoend}
Kanazawa, A., Black, M.J., Jacobs, D.W., Malik, J.: End-to-end recovery of
  human shape and pose. In: {IEEE} Conf. on Computer Vision and Pattern
  Recognition. IEEE Computer Society (2018)

\bibitem{Keyang_2020_ECCV}
Keyang, Z., Bhatnagar, B.L., Pons-Moll, G.: Unsupervised shape and pose
  disentanglement for 3d meshes. In: The European Conference on Computer Vision
  (ECCV) (August 2020)

\bibitem{mh_Pavlakos18}
Kolotouros, N., Pavlakos, G., Black, M.J., Daniilidis, K.: Learning to
  reconstruct 3d human pose and shape via model-fitting in the loop. {IEEE}
  Conf. on Computer Vision and Pattern Recognition  (2019)

\bibitem{mh_Kolotouros19}
Kolotouros, N., Pavlakos, G., Daniilidis, K.: Convolutional mesh regression for
  single-image human shape reconstruction. In: {IEEE} Conference on Computer
  Vision and Pattern Recognition, {CVPR} 2019, Long Beach, CA, USA, June 16-20,
  2019. pp. 4501--4510 (2019)

\bibitem{lazova3dv2019}
Lazova, V., Insafutdinov, E., Pons-Moll, G.: 360-degree textures of people in
  clothing from a single image. In: International Conference on 3D Vision (3DV)
  (sep 2019)

\bibitem{vh_Leroy18}
Leroy, V., Franco, J., Boyer, E.: Shape reconstruction using volume sweeping
  and learned photoconsistency. In: Davis, L.S. (ed.) Computer Vision - {ECCV}
  2018 - 15th European Conference, Munich, Germany, September 8-14, 2018,
  Proceedings, Part {IX}. pp. 796--811. Springer US (2018).
  \doi{10.1007/978-3-030-01240-3\_48}

\bibitem{smpl2015loper}
Loper, M., Mahmood, N., Romero, J., Pons-Moll, G., Black, M.J.: {SMPL}: A
  skinned multi-person linear model. vol.~34, pp. 248:1--248:16. Association
  for Computing Machinery (2015)

\bibitem{LorensenC87mcubes}
Lorensen, W.E., Cline, H.E.: Marching cubes: A high resolution 3d surface
  construction algorithm. In: SIGGRAPH. pp. 163--169. ACM (1987)

\bibitem{i_OccNet19}
Mescheder, L.M., Oechsle, M., Niemeyer, M., Nowozin, S., Geiger, A.: Occupancy
  networks: Learning 3d reconstruction in function space. In: {IEEE} Conference
  on Computer Vision and Pattern Recognition, {CVPR} 2019, Long Beach, CA, USA,
  June 16-20, 2019. pp. 4460--4470 (2019)

\bibitem{i_DeepLevelSet}
Michalkiewicz, M., Pontes, J.K., Jack, D., Baktashmotlagh, M., Eriksson, A.P.:
  Deep level sets: Implicit surface representations for 3d shape inference.
  CoRR  \textbf{abs/1901.06802} (2019), \url{http://arxiv.org/abs/1901.06802}

\bibitem{ih_Dynamic_Fusion}
Newcombe, R.A., Fox, D., Seitz, S.M.: Dynamicfusion: Reconstruction and
  tracking of non-rigid scenes in real-time. In: {IEEE} Conference on Computer
  Vision and Pattern Recognition, {CVPR} 2015, Boston, MA, USA, June 7-12,
  2015. pp. 343--352 (2015). \doi{10.1109/CVPR.2015.7298631}

\bibitem{mh_omran2018neural}
Omran, M., Lassner, C., Pons-Moll, G., Gehler, P., Schiele, B.: Neural body
  fitting: Unifying deep learning and model based human pose and shape
  estimation. In: International Conf. on 3D Vision (2018)

\bibitem{i_DeepSDF}
Park, J.J., Florence, P., Straub, J., Newcombe, R.A., Lovegrove, S.: Deepsdf:
  Learning continuous signed distance functions for shape representation. In:
  {IEEE} Conference on Computer Vision and Pattern Recognition, {CVPR} 2019,
  Long Beach, CA, USA, June 16-20, 2019. pp. 165--174 (2019)

\bibitem{mh_patel2020}
Patel, C., Liao, Z., Pons-Moll, G.: The virtual tailor: Predicting clothing in
  3d as a function of human pose, shape and garment style. In: {IEEE}
  Conference on Computer Vision and Pattern Recognition (CVPR). {IEEE} (Jun
  2020)

\bibitem{ponsmoll2017clothcap}
Pons-Moll, G., Pujades, S., Hu, S., Black, M.: {ClothCap}: Seamless {4D}
  clothing capture and retargeting. ACM Transactions on Graphics
  \textbf{36}(4) (2017)

\bibitem{pons2015dyna}
Pons-Moll, G., Romero, J., Mahmood, N., Black, M.J.: Dyna: a model of dynamic
  human shape in motion. ACM Transactions on Graphics  \textbf{34}, ~120 (2015)

\bibitem{ponsmolMetricForests13}
Pons-Moll, G., Taylor, J., Shotton, J., Hertzmann, A., Fitzgibbon, A.: Metric
  regression forests for human pose estimation. In: British Machine Vision
  Conference (BMVC). BMVA Press (sep 2013)

\bibitem{Pons-Moll_MRFIJCV}
Pons-Moll, G., Taylor, J., Shotton, J., Hertzmann, A., Fitzgibbon, A.: Metric
  regression forests for correspondence estimation. International Journal of
  Computer Vision pp. 1--13 (2015)

\bibitem{popa2017deep}
Popa, A.I., Zanfir, M., Sminchisescu, C.: Deep multitask architecture for
  integrated 2d and 3d human sensing. In: {IEEE} Conf. on Computer Vision and
  Pattern Recognition (2017)

\bibitem{mh_Pumarola}
Pumarola, A., Sanchez, J., Choi, G.P.T., Sanfeliu, A., Moreno{-}Noguer, F.:
  3dpeople: Modeling the geometry of dressed humans. CoRR
  \textbf{abs/1904.04571} (2019)

\bibitem{mh_rhodin2016}
Rhodin, H., Robertini, N., Casas, D., Richardt, C., Seidel, H.P., Theobalt, C.:
  General automatic human shape and motion capture using volumetric contour
  cues. In: European conference on computer vision. pp. 509--526. Springer
  (2016)

\bibitem{MANO:SIGGRAPHASIA:2017}
Romero, J., Tzionas, D., Black, M.J.: Embodied hands: Modeling and capturing
  hands and bodies together. ACM Transactions on Graphics, (Proc. SIGGRAPH
  Asia)  \textbf{36}(6) (Nov 2017)

\bibitem{mh_Rong_2019_ICCV}
Rong, Y., Liu, Z., Li, C., Cao, K., Loy, C.C.: Delving deep into hybrid
  annotations for 3d human recovery in the wild. In: The IEEE International
  Conference on Computer Vision (ICCV) (October 2019)

\bibitem{ih_PiFu}
Saito, S., Huang, Z., Natsume, R., Morishima, S., Kanazawa, A., Li, H.: Pifu:
  Pixel-aligned implicit function for high-resolution clothed human
  digitization. CoRR  \textbf{abs/1905.05172} (2019)

\bibitem{ih_KillingFusion}
Slavcheva, M., Baust, M., Cremers, D., Ilic, S.: Killingfusion: Non-rigid 3d
  reconstruction without correspondences. In: 2017 {IEEE} Conference on
  Computer Vision and Pattern Recognition, {CVPR} 2017, Honolulu, HI, USA, July
  21-26, 2017. pp. 5474--5483 (2017). \doi{10.1109/CVPR.2017.581}

\bibitem{vh_facsimile}
Smith, D., Loper, M., Hu, X., Mavroidis, P., Romero, J.: {FACSIMILE:} fast and
  accurate scans from an image in less than a second. {IEEE} International
  Conference on Computer Vision, {ICCV}  (2019)

\bibitem{mh_stoll}
Stoll, C., Hasler, N., Gall, J., Seidel, H., Theobalt, C.: Fast articulated
  motion tracking using a sums of gaussians body model. In: {IEEE}
  International Conference on Computer Vision, {ICCV} 2011, Barcelona, Spain,
  November 6-13, 2011. pp. 951--958 (2011). \doi{10.1109/ICCV.2011.6126338}

\bibitem{taylor2012vitruvian}
Taylor, J., Shotton, J., Sharp, T., Fitzgibbon, A.: The vitruvian manifold:
  Inferring dense correspondences for one-shot human pose estimation. In: 2012
  IEEE Conference on Computer Vision and Pattern Recognition. pp. 103--110.
  IEEE (2012)

\bibitem{tiwari20sizer}
Tiwari, G., Bhatnagar, B.L., Tung, T., Pons-Moll, G.: Sizer: A dataset and
  model for parsing 3d clothing and learning size sensitive 3d clothing. In:
  European Conference on Computer Vision ({ECCV}). {Springer} (aug 2020)

\bibitem{mh_tung2017self}
Tung, H.Y., Tung, H.W., Yumer, E., Fragkiadaki, K.: Self-supervised learning of
  motion capture. In: Advances in Neural Information Processing Systems. pp.
  5236--5246 (2017)

\bibitem{vh_varol2018bodynet}
Varol, G., Ceylan, D., Russell, B., Yang, J., Yumer, E., Laptev, I., Schmid,
  C.: Bodynet: Volumetric inference of 3d human body shapes. In: Ferrari, V.,
  Hebert, M., Sminchisescu, C., Weiss, Y. (eds.) European Conf. on Computer
  Vision. Springer International Publishing (2018)

\bibitem{wei2016dense}
Wei, L., Huang, Q., Ceylan, D., Vouga, E., Li, H.: Dense human body
  correspondences using convolutional networks. In: Computer Vision and Pattern
  Recognition (CVPR) (2016)

\bibitem{mh_Xiang_2019_CVPR}
Xiang, D., Joo, H., Sheikh, Y.: Monocular total capture: Posing face, body, and
  hands in the wild. In: The IEEE Conference on Computer Vision and Pattern
  Recognition (CVPR) (June 2019)

\bibitem{xu2020ghum}
Xu, H., Bazavan, E.G., Zanfir, A., Freeman, W.T., Sukthankar, R., Sminchisescu,
  C.: Ghum \& ghuml: Generative 3d human shape and articulated pose models. In:
  CVPR (2020)

\bibitem{yang2016estimation}
Yang, J., Franco, J.S., H{\'e}troy-Wheeler, F., Wuhrer, S.: Estimation of human
  body shape in motion with wide clothing. In: Leibe, B., Matas, J., Sebe, N.,
  Welling, M. (eds.) European Conference on Computer Vision. Springer
  International Publishing (2016)

\bibitem{ih_DoubleFusion}
Yu, T., Zheng, Z., Guo, K., Zhao, J., Dai, Q., Li, H., Pons{-}Moll, G., Liu,
  Y.: Doublefusion: Real-time capture of human performances with inner body
  shapes from a single depth sensor. In: 2018 {IEEE} Conference on Computer
  Vision and Pattern Recognition, {CVPR} 2018, Salt Lake City, UT, USA, June
  18-22, 2018. pp. 7287--7296 (2018). \doi{10.1109/CVPR.2018.00761}

\bibitem{zanfir2020weakly}
Zanfir, A., Bazavan, E.G., Xu, H., Freeman, B., Sukthankar, R., Sminchisescu,
  C.: Weakly supervised 3d human pose and shape reconstruction with normalizing
  flows. European Conf. on Computer Vision  (2020)

\bibitem{mh_zanfir2018monocular}
Zanfir, A., Marinoiu, E., Sminchisescu, C.: Monocular 3d pose and shape
  estimation of multiple people in natural scenes--the importance of multiple
  scene constraints. In: Proceedings of the IEEE Conference on Computer Vision
  and Pattern Recognition. pp. 2148--2157 (2018)

\bibitem{zanfir2018monocular}
Zanfir, A., Marinoiu, E., Sminchisescu, C.: Monocular 3d pose and shape
  estimation of multiple people in natural scenes--the importance of multiple
  scene constraints. In: Proceedings of the IEEE Conference on Computer Vision
  and Pattern Recognition. pp. 2148--2157 (2018)

\bibitem{zanfir_nips2018}
Zanfir, A., Marinoiu, E., Zanfir, M., Popa, A.I., Sminchisescu, C.: Deep
  network for the integrated 3d sensing of multiple people in natural images.
  In: NIPS (2018)

\bibitem{zhang2017shapeundercloth}
Zhang, C., Pujades, S., Black, M., Pons-Moll, G.: Detailed, accurate, human
  shape estimation from clothed {3D} scan sequences. In: {IEEE} Conf. on
  Computer Vision and Pattern Recognition (2017)

\bibitem{vh_DeepHumans}
Zheng, Z., Yu, T., Wei, Y., Dai, Q., Liu, Y.: Deephuman: 3d human
  reconstruction from a single image. The IEEE International Conference on
  Computer Vision (ICCV)  (2019)

\end{thebibliography}
\end{document}